\documentclass[runningheads]{llncs}
\usepackage[T1]{fontenc}
\usepackage{hyperref}
\usepackage{booktabs}
\usepackage{xcolor}
\usepackage{multirow}
\usepackage{listings}
\usepackage{tcolorbox}
\usepackage{graphicx}
\begin{document}
\title{When Low CER is Not Enough: An Analysis of Hallucinations in Vision-Language OCR Systems on Historical Uruguayan Documents}
%
%
\author{Marina Gardella\inst{1}$^{\dagger}$\orcidID{0000-0003-2465-2014} \and 
Camilo Mariño\inst{1,2}$^{\dagger}$\orcidID{0009-0005-7094-9371} \and 
Diego Belzarena\inst{1,2}$^{\dagger}$\orcidID{0009-0006-0535-7246} \and 
Ignacio Ramírez\inst{2}\orcidID{0000-0003-2954-9040} \and 
Gregory Randall\inst{2}\orcidID{0000-0001-7911-2977} \and 
Jean-Michel Morel\inst{3}\orcidID{0000-0002-6108-897X}
}
\authorrunning{M. Gardella et al.
}
\institute{Université Paris-Saclay, ENS Paris-Saclay, CNRS, Centre Borelli, France \and
Facultad de Ingeniería, Universidad de la República, Montevideo, Uruguay \and
Lingnan University, Hong Kong
}
\maketitle              
\begingroup
\renewcommand{\thefootnote}{\dag}
\footnotetext{These authors contributed equally.}
\endgroup
\begin{abstract}
Optical Character Recognition (OCR) is a key component in the digitization of historical archives. Recently, Vision-Language Models (VLMs) have emerged as strong alternatives to traditional OCR systems, achieving state-of-the-art performance on standard benchmarks. However, their suitability for archival transcription remains insufficiently understood. In this work, we benchmark traditional OCR systems and VLM-based approaches on the Berrutti dataset, a challenging collection of Uruguayan dictatorship-era documents derived from microfilm scans. While VLMs consistently outperform traditional methods in terms of Character Error Rate (CER) and Word Error Rate (WER), we show that these improvements hide a more complex picture. Through a detailed qualitative analysis, we uncover systematic failure modes that are invisible to standard metrics, including orthographic normalization, spurious content generation, and semantic substitutions that preserve fluency while altering meaning. Errors affecting named entities are particularly critical, as they can introduce substantial semantic distortions with minimal impact on CER and WER. These findings reveal a critical gap between quantitative OCR performance and transcription fidelity in real-world archival settings, and highlight the need for evaluation frameworks that go beyond character-level accuracy to capture the semantic reliability of generated transcriptions.
\keywords{Optical Character Recognition \and Vision-Language Models \and Historical Document Analysis \and Evaluation Metrics \and Hallucinations}

\end{abstract}

\section{Introduction}
\label{sec:intro}

The digitization of historical archives relies on Optical Character Recognition (OCR) technologies to convert scanned documents into searchable and analyzable text. In recent years, Vision-Language Models (VLMs) have emerged as a powerful alternative to traditional OCR systems, achieving remarkable performance across a wide range of document understanding tasks. By combining visual perception with large-scale language modeling capabilities, these systems can leverage contextual information to recover degraded or ambiguous text that would often challenge conventional OCR pipelines.

OCR performance is typically assessed using character-level and word-level metrics, such as Character Error Rate (CER) and Word Error Rate (WER). These measures have become standard for comparing transcription systems and are widely used in both modern and historical document analysis. However, they implicitly treat all transcription errors similarly, regardless of their semantic importance. This assumption becomes problematic in the context of VLM-based transcription, as these models can generate linguistically plausible text that is not fully supported by visual evidence. Such behavior can manifest itself as contextual completions, orthographic normalization, semantic substitutions, or outright spurious additions. Names and dates-less determined by contextual information-are more prone to these errors. Although these errors may have limited impact on CER or WER, they can alter the factual content of a document and compromise its value for archival, historical, or scholarly analysis. 

For historical documents, transcription fidelity is often more important than linguistic plausibility. Historians, archivists, and researchers seek accurate representations of the original source~\cite{challenge2026icdar}. A transcription that silently corrects, completes, or reinterprets the source text may therefore be less desirable than a transcription that contains minor recognition errors. Previous work has shown that OCR errors can have markedly different effects on downstream retrieval and information extraction tasks depending on the information they affect, with errors involving key entities often having a disproportionately large impact despite contributing little to aggregate CER or WER values~\cite{BeyondCERWER}.

In this work, we investigate these issues through a benchmark of traditional OCR systems and recent VLM-based transcription models on \href{https://github.com/camilomarino/ocr_berrutti_dataset}{Berrutti}~\cite{belzarena2025improving}, a dataset of historical Uruguayan documents. We first compare systems using conventional OCR metrics and show that VLMs consistently achieve lower CER and WER than traditional approaches. We then perform a qualitative analysis of their transcription errors and identify failure modes that are not adequately reflected by standard evaluation metrics. Our findings reveal that the superior quantitative performance of VLMs can coexist with visually ungrounded and semantically significant errors, highlighting the need for evaluation protocols that go beyond character-level accuracy when assessing OCR systems.

The Berrutti dataset provides a particularly relevant setting for this analysis. Beyond the challenges commonly associated with historical archives, the collection contains documents related to human rights investigations and transitional justice processes in which faithful transcription is essential to preserve documentary evidence and support subsequent retrieval, analysis, and information extraction. In such contexts, seemingly minor transcription alterations can affect the interpretation of historically significant information, making transcription reliability as important as transcription accuracy.

\section{Related Works}
\label{sec:works}

\subsection{Traditional OCR and Historical Documents}

Historically, OCR pipelines relied on heavy preprocessing, explicit character segmentation, and classifiers based on hand-crafted features. In recent years, however, the field has shifted toward deep learning models. By learning representations directly from human-annotated corpora, these models can transcribe text end-to-end, significantly reducing the need for manual image normalization and explicit character segmentation~\cite{ocropus,ctc,puigcerver,crnn,tesseract}. Despite substantial improvements in recognition accuracy, most of these systems remain composed of multiple quasi-independent stages, including layout analysis, text detection, and text recognition~\cite{ppocrv5,doctr2021}, making them susceptible to error propagation. In addition, despite their promises, these types of models often exhibit limited generalization across domains, requiring task-specific adaptation or fine-tuning when confronted with changes in language, typography, writing style, document degradation or acquisition conditions~\cite{barrere,kang2020,pippi2023,simon2026}.

These challenges become particularly pronounced in historical collections, where physical degradation, irregular layouts, handwriting, 
character forms, and historical language variants can significantly affect transcription quality. Consequently, numerous specialized OCR architectures and dedicated transcription platforms have been proposed for historical documents~\cite{escriptorium,transkribus,kraken,ocr-d,ocr4all,calamari}. Nevertheless, robust generalization across collections remains an open problem, particularly when annotated training data is scarce~\cite{low-resource,challenge2026icdar}.

\subsection{Vision-Language Models and the Challenge of Hallucinations}

Recent advances in VLMs, ranging from general-purpose multimodal models prompted for document understanding~\cite{qwen3_6_alibaba,google_gemma4_docs,yu2025minicpmv45} to specialized end-to-end document VLMs~\cite{paddleocrvl15multitask09bvlm,chandra,qianfan,duan2026glmocrtechnicalreport,dotsOCR,paruchuri2025surya,hunyuanvisionteam2025hunyuanocrtechnicalreport,mineru25p,wei2025deepseek,dotsMOCR}, have introduced a new paradigm for document transcription.
Unlike conventional OCR pipelines, VLM-based approaches can formulate transcription as a direct image-to-text generation task, using large-scale multimodal pretraining to combine visual evidence with linguistic context. This ability to exploit contextual information has enabled state-of-the-art performance on numerous OCR and document understanding benchmarks and has often improved cross-domain robustness~\cite{ocrbench,olmbench}. 
However, many benchmarks used to evaluate modern OCR and VLM systems are dominated by contemporary standardized documents characterized by mostly clean scans and modern language usage~\cite{ocrbench,omnidocbench}. Consequently, benchmark improvements may partly reflect a model's ability to exploit linguistic regularities rather than its capacity to faithfully recover visually ambiguous text. 
When applied to historical collections with severe visual degradation, this reliance on linguistic priors may become a liability~\cite{challenge2026icdar,zhang2026democratizingmedievalenglishlegal}.

This liability manifests itself as hallucination-related errors. Although the term \textit{hallucination} is commonly used in the context of LLMs to describe responses that are false or misleading, yet presented as certain or factual, its meaning in OCR remains less clear.
Recent comparative studies, particularly in the context of historical manuscripts, have increasingly acknowledged this particular behavior of current methods. These studies either complement standard metrics with qualitative analyses of hallucination-like behaviors~\cite{challenge2026icdar,churro,zhang2026democratizingmedievalenglishlegal}, 
or explicitly report hallucinations as a separate evaluation category alongside CER and WER~\cite{metatr}. Although these efforts reflect a growing recognition that character-level accuracy alone may not fully capture transcription reliability, they often lack standardized detection methodologies~\cite{metatr}. 

Given that there is still no broad consensus on how to identify or measure these phenomena, comparing systems remains difficult. To navigate this ambiguity, we adopt a working definition for the scope of this paper: we classify an error as a hallucination if the generated text lacks direct visual evidence in the source image or actively contradicts it, despite being linguistically plausible. Under this definition, a visual misclassification (such as mistaking an 'e' for a 'c' due to image degradation) remains a traditional recognition error, whereas generating a contextually appropriate but visually absent word constitutes a hallucination.

\subsection{Downstream Impacts and the Need for Qualitative Analysis}

Understanding the qualitative nature of these errors is particularly relevant because OCR transcriptions serve as the primary input to larger document-processing pipelines. Previous studies have extensively documented how transcription errors degrade the performance of downstream NLP tasks~\cite{assessingOCRNLP}. In particular, research on named entity processing has shown that errors disproportionately disrupt document access and organization when they affect key entities~\cite{ocrNERlinking}. Still, studies on post-OCR correction have demonstrated that mitigating these downstream effects is possible, provided the initial transcription errors fall within certain bounds and exhibit predictable patterns~\cite{whenPostCorrectionNER}. 
However, this existing body of work is largely grounded in the error characteristics of traditional OCR systems, in which errors typically arise from visual recognition failures and manifest as character substitutions, insertions, deletions, or segmentation errors~\cite{ocrNERlinking}. 

Consequently, there remains little understanding of how the qualitatively different behaviors observed in modern VLM-based OCR systems (e.g., semantic substitutions and spurious additions) may affect the reliability of transcriptions. Although the present work does not directly measure NLP performance, systematically categorizing these specific error modes is a practical first step toward understanding their real-world impact. This is especially relevant for high-stakes domains such as human-rights archives used in transitional justice~\cite{lessa2013,TransitionalJustice}, where digitized documents are consulted to reconstruct historical events. In these contexts, transcription errors do not merely affect character-level accuracy, but can compromise the faithful preservation of documentary evidence.

\section{Dataset}
\label{sec:dataset}

\setlength{\fboxrule}{0.5pt}
\setlength{\fboxsep}{0pt}
\begin{figure}[ht]
    \centering
    \fbox{\includegraphics[height=0.4\linewidth]{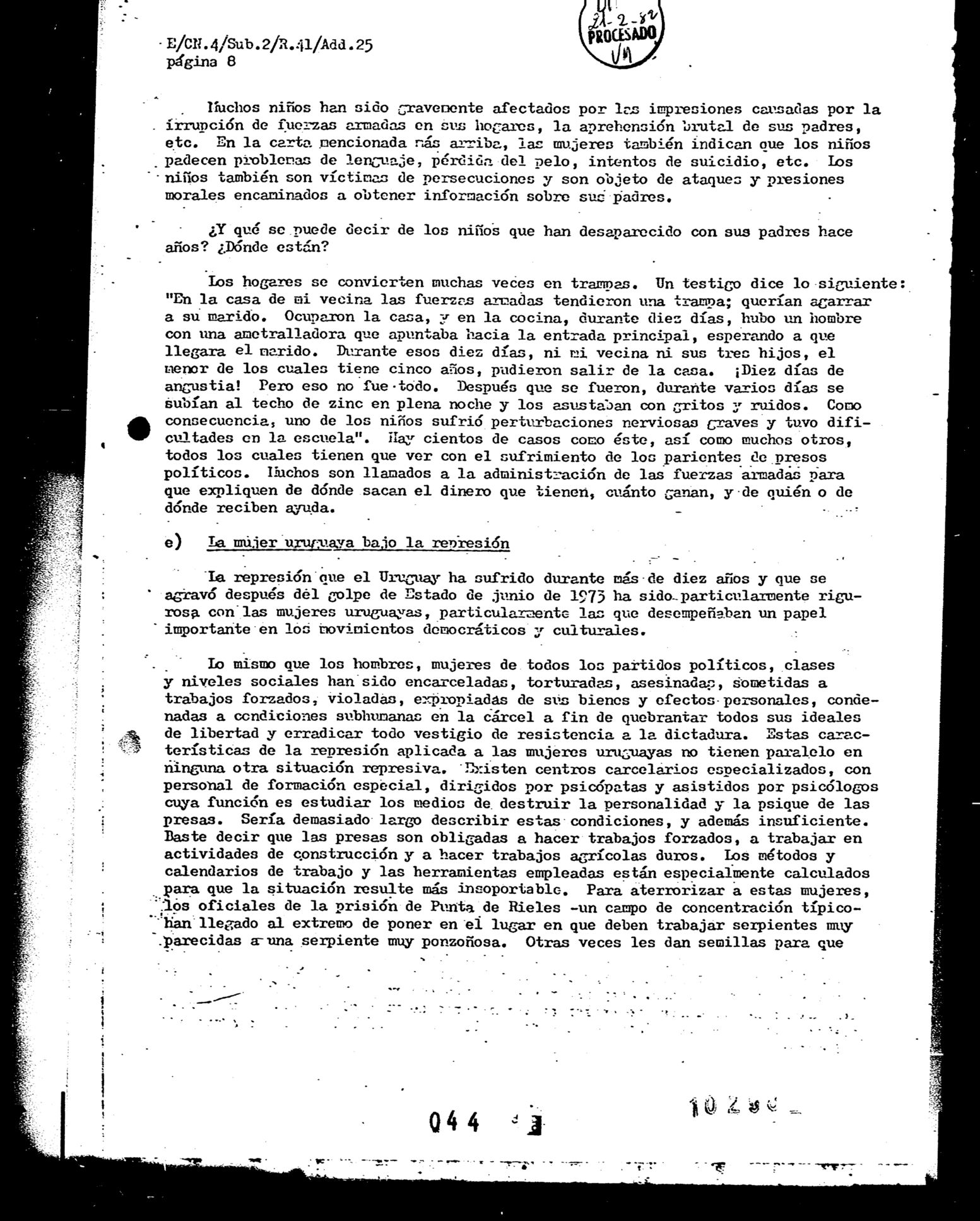}}
    \fbox{\includegraphics[height=0.4\linewidth]{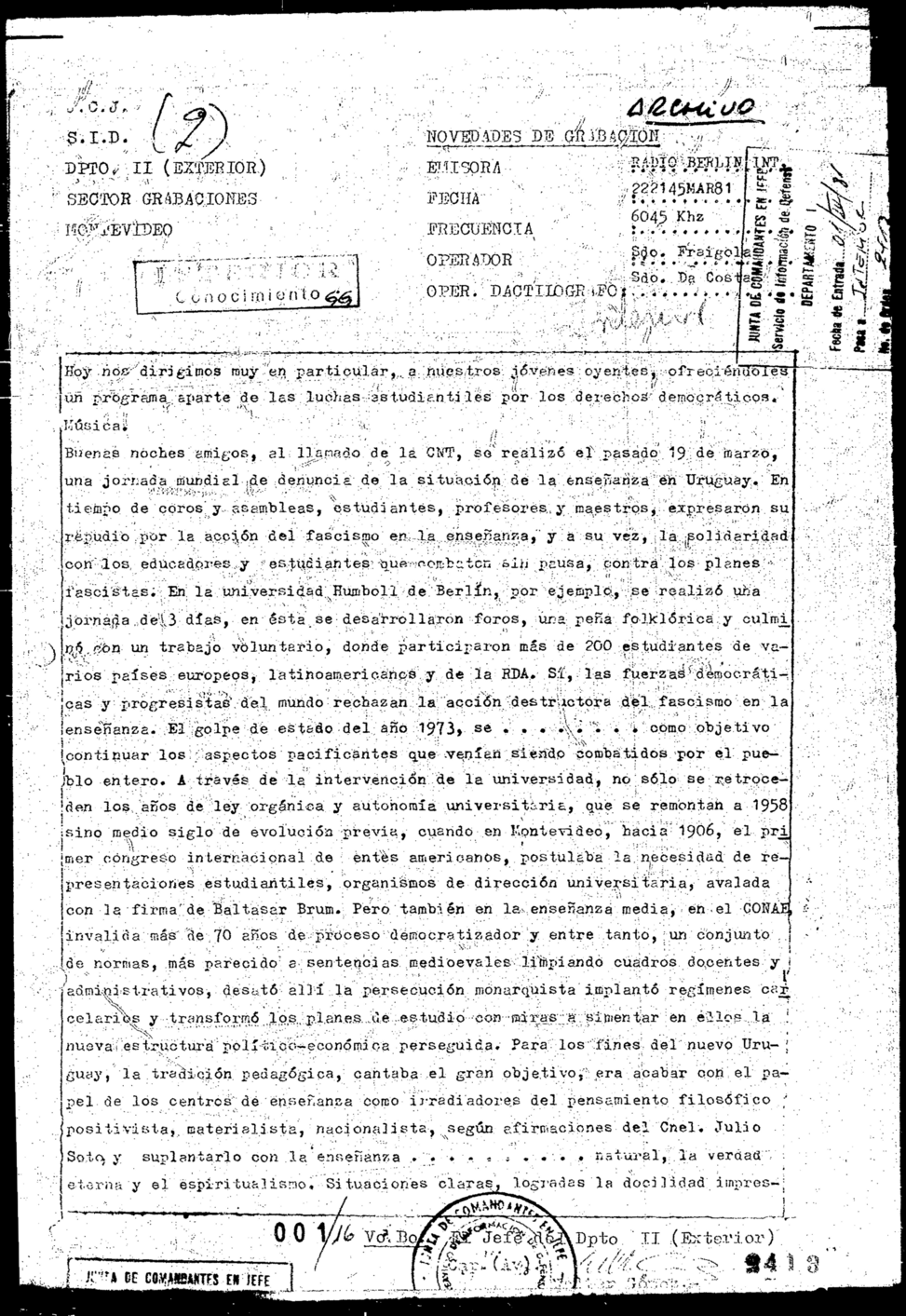}}
    \fbox{\includegraphics[height=0.4\linewidth]{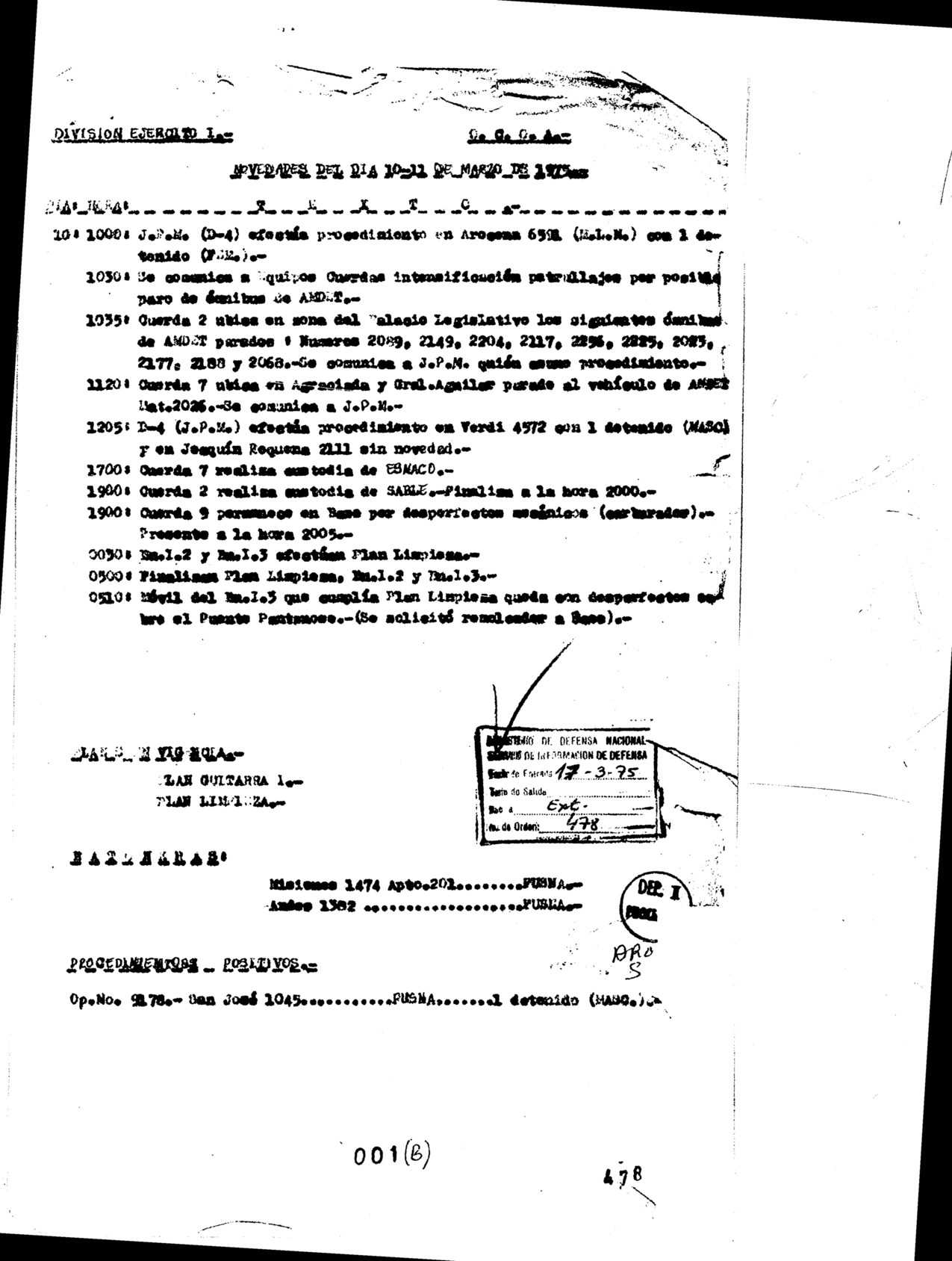}}
    \caption{Representative samples from the Berrutti dataset, illustrating varying levels of document degradation commonly encountered in binary microfilm scans.}
    \label{fig:berrutti-samples}
\end{figure}

The Berrutti dataset was released in 2025. It is derived from the Berrutti Archive, a large microfilm collection containing over 3 million pages mainly produced during the Uruguayan military dictatorship (1973--1985). The archive consists primarily of typewritten reports and other documents produced by government agencies, often accompanied by handwritten annotations.

Beyond its documentary value, the archive constitutes a key resource for historical research, human rights investigations, and transitional justice processes in Uruguay. Previous studies have highlighted the importance of repressive archives as sources for reconstructing state violence, identifying victims, and supporting truth and justice initiatives~\cite{caetano2011,caetano2017,lessa2013,lessa2022}. As a consequence, accurate document transcription is not only an OCR task but also a prerequisite for reliable search, indexing, information extraction, and downstream historical analyses.

The images were digitized directly as binary (bi-level) images from the microfilm sources. Consequently, the dataset contains real degradation artifacts, including heavy non-additive noise, overlapping visual elements, and scanning defects, which make text recognition particularly challenging~\cite{etcheverry2021}. Fig.~\ref{fig:berrutti-samples} presents some examples from the dataset.

For research purposes, the Uruguayan \textit{Institución Nacional de Derechos Humanos y Defensoría del Pueblo} authorized the public release of a curated subset comprising 175 document images, which constitute the dataset used in this work. The selected documents were manually reviewed to ensure compliance with legal and ethical requirements and were chosen to be representative of the broader collection while excluding sensitive personal information. 
The collection is available at \href{https://github.com/camilomarino/ocr_berrutti_dataset}{https://github.com/camilomarino/ocr\_berrutti\_dataset}.

\section{Quantitative Evaluation} 
\subsection{Evaluated methods}
\label{sec:methods}

We evaluated a diverse set of open-source OCR systems, which we categorized into traditional OCR systems, specialized OCR-oriented VLMs, and general-purpose VLMs. Traditional 
OCR systems include PP-OCRv5~\cite{ppocrv5}, EasyOCR~\cite{easyocr}, DocTR~\cite{doctr2021} and Tesseract~\cite{tesseract}. Specialized VLMs comprise recent models explicitly designed for document understanding and text recognition, including Surya-OCR~\cite{paruchuri2025surya}, Hunyuan-OCR~\cite{hunyuanvisionteam2025hunyuanocrtechnicalreport}, MinerU2.5-Pro~\cite{mineru25p}, Chandra-OCR 2~\cite{chandra}, GLM-OCR~\cite{duan2026glmocrtechnicalreport}, Qianfan-OCR~\cite{qianfan}, Dots-MOCR~\cite{dotsMOCR}, Dots-OCR~\cite{dotsOCR}, PaddleOCR-VL 1.5~\cite{paddleocrvl15multitask09bvlm}, and DeepSeek-OCR2~\cite{wei2025deepseek}. Finally, we evaluated general-purpose VLMs, namely Qwen 3.6~\cite{qwen3_6_alibaba}, Gemma 4~\cite{google_gemma4_docs}, and MiniCPM-V 4.5~\cite{yu2025minicpmv45}.

The selected systems cover a broad spectrum of OCR approaches, ranging from traditional recognition pipelines to large generative multimodal models. Due to the sensitive nature of the Berrutti archive and its associated privacy requirements, our evaluation was restricted to models that can be executed locally without transmitting document images to external services, even though the subset used here has been publicly released under appropriate authorization.

\subsection{Evaluation protocol}
\label{sec:evaluation}

For prompt-based models, we used the same OCR-oriented instruction whenever free-form prompting was available. For systems with a fixed OCR prompt, we used the instruction provided without modification. The exact prompt used in the former case is shown below:

\begin{tcolorbox}[colback=gray!5, colframe=gray!50, title=OCR-oriented Prompt]
\footnotesize
OCR this scanned historical Uruguayan document page. It is a typewritten document from the Uruguayan dictatorship period.

Return only plain text in reading order. Preserve original line breaks, spelling, accents, capitalization, punctuation, abbreviations, dates, numbers, and hyphenation.

Read top-to-bottom. In multi-column or table-like regions, use the natural human order: columns left-to-right; rows top-to-bottom and cells left-to-right.

Do not explain. Do not summarize. Do not use markdown. Do not correct or expand abbreviations. Do not invent missing text. Use [ilegible] only for unreadable text.
\end{tcolorbox}

All methods were evaluated on the Berrutti dataset, described in Sect.~\ref{sec:dataset}. For each document, model output was compared with the ground-truth transcription using Character Error Rate (CER) and Word Error Rate (WER), computed using Levenshtein alignment.

Before evaluation, both reference and predicted texts were processed with the same normalization pipeline. This step removed formatting artifacts sometimes added by document-parsing models, such as Markdown headings, code fences, links, table separators, HTML tags, and image references. Normalization preserved case, accents, punctuation, hyphenation, and line breaks. It only converted Windows-style line endings, collapsed repeated horizontal whitespace into a single space, and trimmed the leading or trailing whitespace on each line. This avoids penalizing minor spacing differences while keeping the textual properties relevant to our analysis.

To account for the large variability in document difficulty, we report both the mean and median error rates in the dataset. While the mean incorporates the contribution of all documents and is sensitive to outliers, the median is less influenced by a small number of high-error cases, providing a complementary view of model performance across the dataset.


\subsection{Results}
\label{sec:results}

Tab.~\ref{tab:berrutti-ocr-ranked} summarizes the quantitative results. General-purpose VLMs achieve the best median performance, with Qwen 3.6 in reasoning mode obtaining the lowest median CER (1.72\%) and median WER (5.01\%). Specialized VLMs also perform strongly, with Surya-OCR achieving the lowest mean CER (8.87\%). In contrast, traditional OCR systems exhibit substantially higher error rates.

\setlength{\tabcolsep}{7pt}
\begin{table}[t]
\centering
\small
\begin{tabular}{l l cccc}
\toprule
Model & Type & \multicolumn{2}{c}{CER} & \multicolumn{2}{c}{WER} \\
\cmidrule(lr){3-4} \cmidrule(lr){5-6}
 &  & Mean & Median & Mean & Median \\
\midrule

Qwen 3.6$^\star$ (35b)  & General VLM & 11.17\% & \textbf{1.72\%} & \textbf{15.12\%} & \textbf{5.01\%} \\
Qwen 3.6 (35b) & General VLM & 9.19\% & 2.01\% & 18.58\% & 5.54\% \\
Gemma 4 (31b) & General VLM & 10.06\% & 3.27\% & 24.29\% & 9.40\% \\
Surya-OCR & Specialized VLM & \textbf{8.87\%} & 4.00\% & 19.13\% & 11.38\% \\
Hunyuan-OCR & Specialized VLM & 12.46\% & 4.07\% & 27.57\% & 13.80\% \\
MinerU2.5-Pro & Specialized VLM & 9.38\% & 4.19\% & 18.78\% & 11.82\% \\
Chandra-OCR 2 & Specialized VLM & 13.19\% & 4.38\% & 22.86\% & 10.45\% \\
GLM-OCR & Specialized VLM & 10.68\% & 4.43\% & 21.60\% & 12.43\% \\
Qianfan-OCR & Specialized VLM & 23.75\% & 4.45\% & 37.69\% & 12.77\% \\
Dots-MOCR & Specialized VLM & 10.36\% & 4.50\% & 22.33\% & 12.50\% \\
MiniCPM-V 4.5  & General VLM & 9.26\% & 5.24\% & 17.79\% & 14.43\% \\
Dots-OCR & Specialized VLM & 19.82\% & 5.54\% & 27.87\% & 14.21\% \\
DocTR & Traditional OCR & 10.23\% & 5.74\% & 31.43\% & 25.15\% \\
PaddleOCR-VL 1.5 & Specialized VLM & 12.06\% & 5.97\% & 21.17\% & 14.80\% \\
DeepSeek-OCR2 & Specialized VLM & 22.45\% & 6.20\% & 43.65\% & 16.14\% \\
PP-OCRv5 & Traditional OCR & 10.87\% & 8.18\% & 27.33\% & 22.42\% \\
Tesseract & Traditional OCR & 20.92\% & 11.86\% & 41.21\% & 29.63\% \\
EasyOCR & Traditional OCR & 25.41\% & 20.70\% & 59.47\% & 62.28\% \\
\bottomrule
\end{tabular}
\caption{OCR benchmark on the Berrutti dataset. Best results per metric are highlighted in bold. Models are categorized as general VLMs, specialized VLMs, or traditional OCR systems. The $\star$ in Qwen 3.6 indicates reasoning mode.}
\label{tab:berrutti-ocr-ranked}
\end{table}

A notable observation is the large difference between mean and median error rates for several VLM-based systems. For example, Qwen 3.6 in reasoning mode (Qwen 3.6$^\star$) achieves a median CER of 1.72\%, while its mean CER reaches 11.17\%. Similar behavior can be observed for other high-performing VLMs. This suggests that although these models produce highly accurate transcriptions for most documents, they occasionally fail, generating errors that disproportionately affect aggregate metrics.


In general, the results in Tab.~\ref{tab:berrutti-ocr-ranked} indicate that VLM-based approaches substantially outperform traditional OCR systems according to the standard metrics used in the literature. However, as we shall discuss in the following section, a closer inspection of the generated transcriptions reveals qualitative differences in the types of error these models produce.

\section{Qualitative Error Analysis}
We identified four recurrent failure modes: incomplete transcriptions, orthographic normalization, spurious additions, and semantic substitutions and insertions. The last two, in particular, arise from the generative nature of VLMs and may alter the meaning of the source document while having only a limited impact on standard OCR metrics.

\subsection{Incomplete Transcription}

 \begin{figure}[t!]
     \centering
\includegraphics[width=\linewidth]{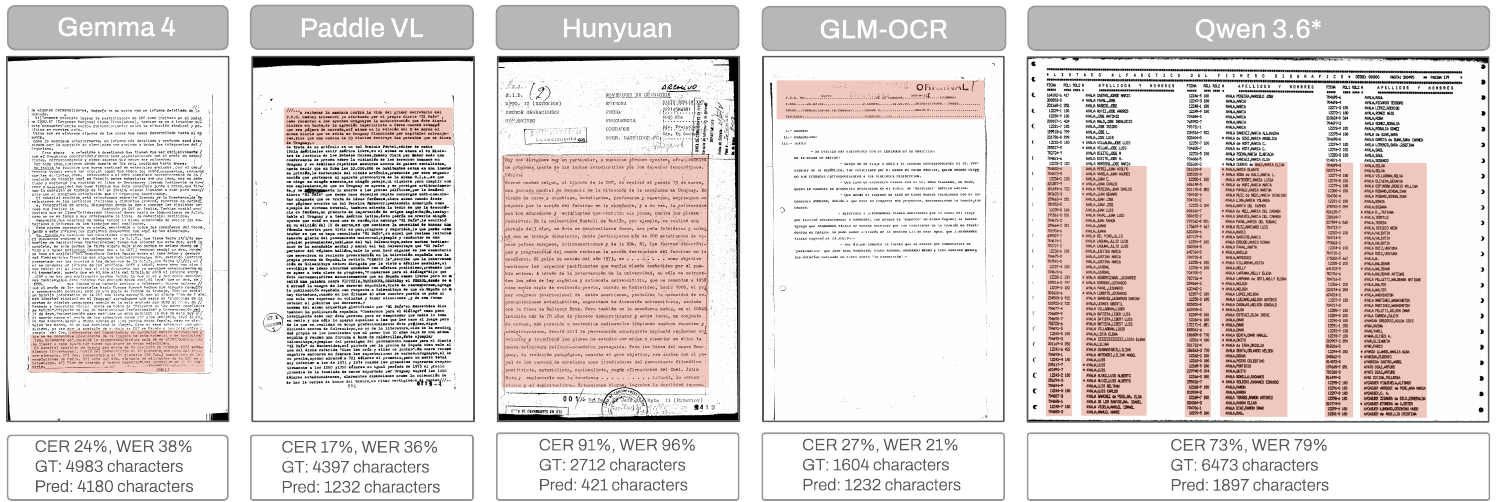}
    \caption{Examples of incomplete 
    transcriptions. The red overlay highlights regions of the document not covered by each model's prediction.}
    \label{fig:incomplete}
\end{figure}

In some cases, the models produce only partial transcriptions, ignoring large portions of the text. These incomplete transcriptions can arise from different types of generation failures. The models may omit content at the beginning, at the end, or in intermediate regions of the document, as depicted in Fig.~\ref{fig:incomplete}.

\begin{figure}[t!]
\centering
\includegraphics[width=0.9\linewidth]{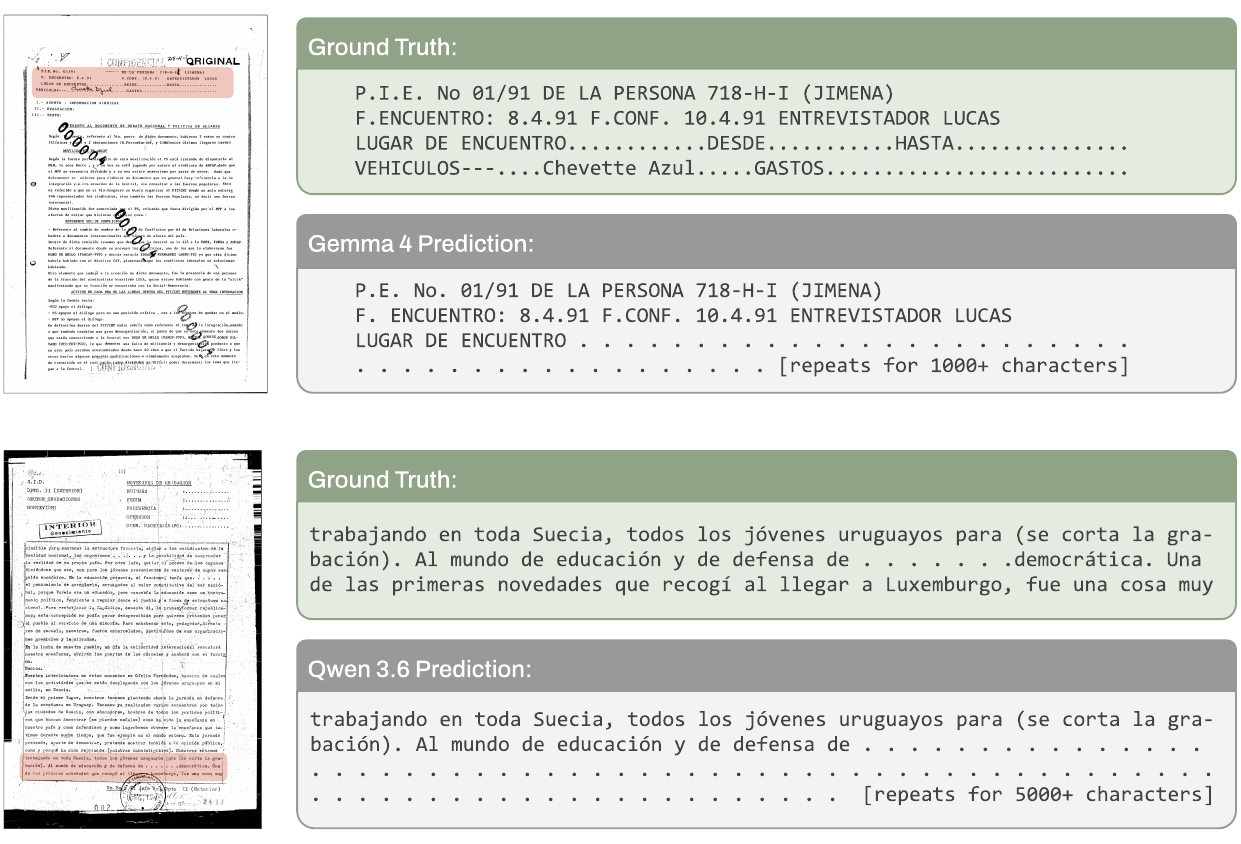}
\caption{Examples of generation collapse. Left: document images, with the regions where the failure begins highlighted in red. Right: ground-truth text for those regions and the corresponding model predictions. Models initially transcribe correctly but later collapse into repetitive token loops.}
\label{fig:degenerate}
\end{figure}

In other cases, transcription is interrupted while generation continues, resulting in repetitive or degenerate output. Fig.~\ref{fig:degenerate} illustrates this behavior, where the model stops producing meaningful transcription, yet continues generating tokens. Interestingly, this behavior often occurs when the document contains long sequences of dots or repetitive punctuation patterns.

In both cases, incomplete transcriptions are generally reflected in standard metrics such as CER and WER. However, unlike character-level recognition errors, these omissions result in the loss of potentially relevant information. 



\subsection{Orthographic Normalization}
Rather than faithfully reproducing the text as it appears in the document, some VLMs implicitly correct spelling, punctuation, or capitalization. Although these modifications often improve linguistic quality and may even reduce perceived reading difficulty, they compromise the objective of producing a faithful transcription of the original source.

\begin{figure}[t!]
\centering
\includegraphics[width=0.9\linewidth]{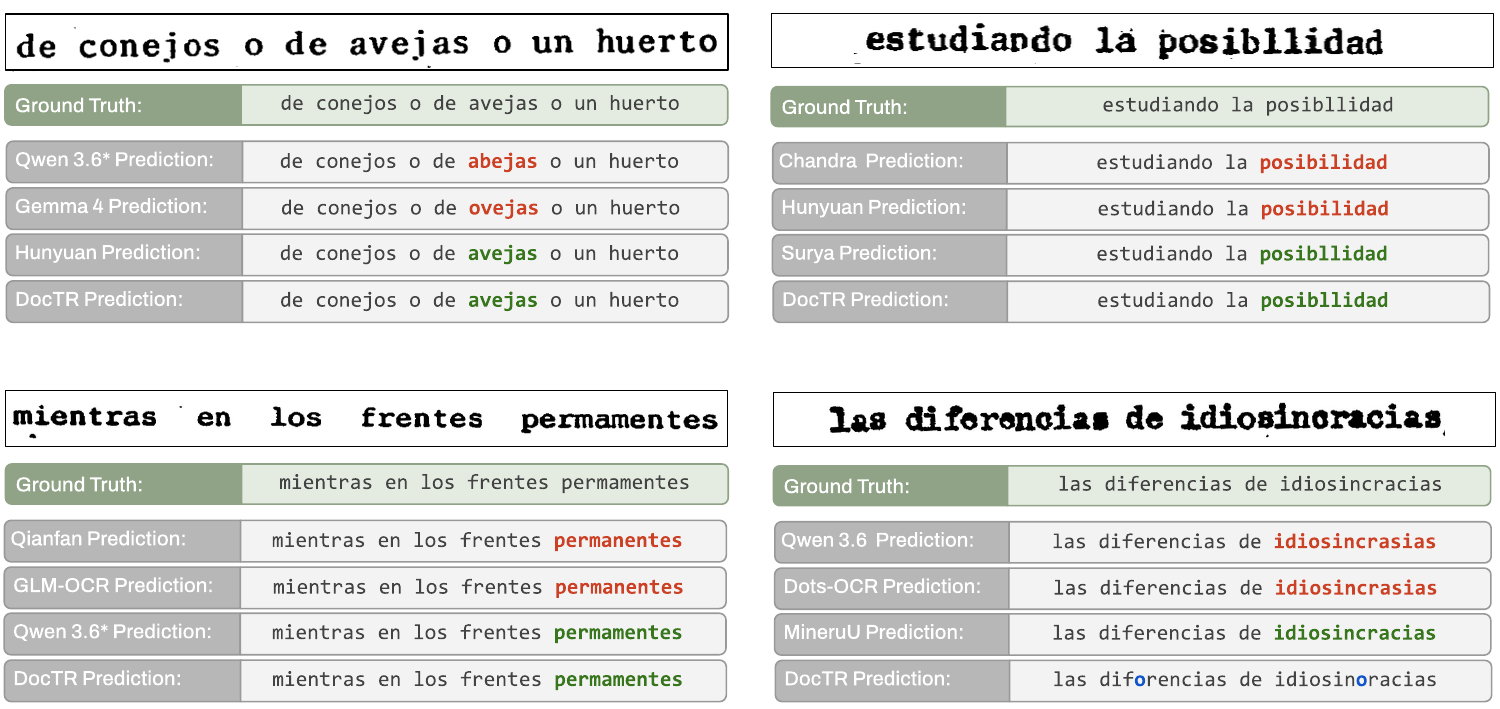}
\caption{Examples of orthographic mistakes and their corresponding transcriptions produced by VLMs and DocTR. Corrected words are highlighted in red while faithful transcriptions are highlighted in green. VLMs may correct spelling errors inconsistently, whereas DocTR preserves the original spelling but may still produce recognition errors, which are highlighted in blue.}
\label{fig:orthographic}
\end{figure}

Fig.~\ref{fig:orthographic} presents four examples of orthographic mistakes found in the original documents, together with four corresponding transcriptions for each example: three produced by VLMs and one by DocTR. Interestingly, when faced with spelling errors, VLMs do not consistently preserve or correct them. A model may correct certain errors while leaving others unchanged. For example, Qwen 3.6$^\star$ corrects \textit{avejas} but preserves \textit{permamentes}. In contrast, DocTR does not appear to perform an orthographic correction; instead, it attempts to reproduce the text as written. However, the absence of orthographic correction should not be interpreted as evidence of higher fidelity, since recognition errors can still alter the original content, as illustrated in the last example.

\begin{figure}[t!]
\centering
\includegraphics[width=\linewidth]{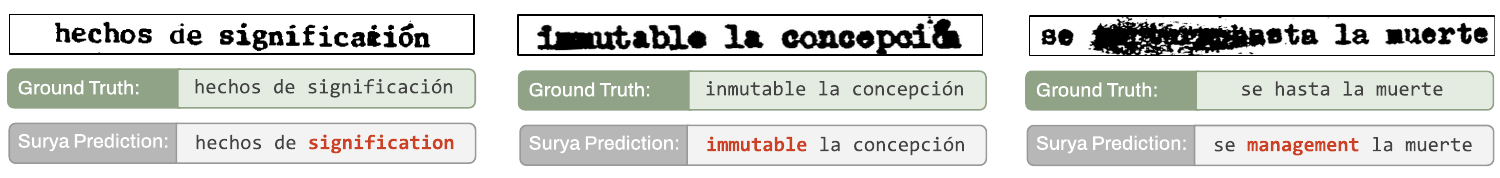}
\caption{Examples of Surya resolving visual ambiguities in favor of English words, highlighted in red.}
\label{fig:surya}
\end{figure}

In our analysis, we observed that Surya performs fewer orthographic corrections than the other VLMs, often reproducing the original spelling more faithfully. Nevertheless, when the visual evidence is ambiguous, the model frequently resolves the uncertainty in favor of English words rather than Spanish ones. This behavior may indicate a stronger prior toward English-language text, which could explain its lower tendency to perform orthographic corrections in Spanish documents. Examples of this phenomenon are shown in Fig.~\ref{fig:surya}.

\subsection{Spurious Additions}

\begin{figure}[t!]
\centering
\includegraphics[width=\linewidth]{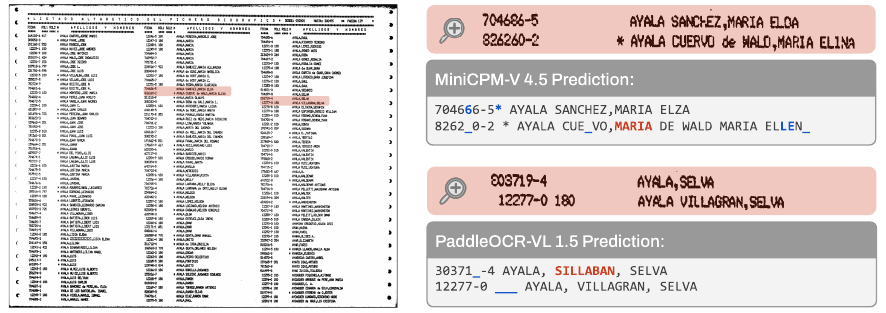}
\caption{Examples of failure cases in names lists, a high-value document type. VLMs introduce additional names (highlighted in red) not present in the original document, thereby compromising named entity integrity. Other errors are highlighted in blue.}
\label{fig:names}
\end{figure}

Spurious additions occur when models generate words, phrases, or fragments that cannot be aligned with any region of the input document. Unlike standard OCR errors, these additions introduce new content into the transcription rather than misrecognizing existing content. Such errors are particularly concerning, as they may appear plausible to a human reader yet lack any basis in the source document, potentially leading to incorrect interpretations of archival records.

\begin{figure}[t!]
\centering
\includegraphics[width=\linewidth]{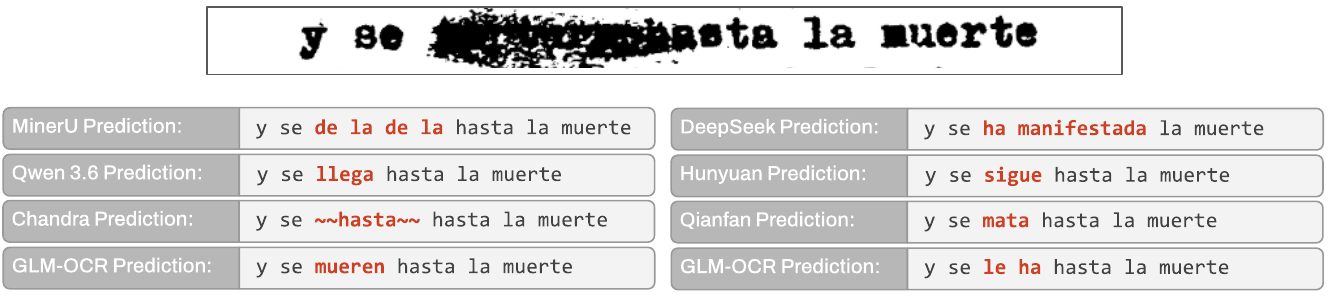}
\caption{Outputs of different VLMs on a region containing no recoverable textual information. Spurious additions are highlighted in red. Models generate plausible words rather than marking the region as unreadable, illustrating a common failure mode.}
\label{fig:muerte}
\end{figure}

Highly degraded regions containing no recoverable textual information appear to be a particularly strong trigger for this behavior. One such example is shown in the third column of Fig.~\ref{fig:surya}, although this phenomenon is by no means limited to that model. Fig.~\ref{fig:muerte} presents the outputs of several other VLMs on the same document region, revealing that most of them generate plausible text despite the absence of any readable content.

Notably, this behavior arises even if the models are explicitly instructed to use the token \texttt{[illegible]} for unreadable text. Rather than expressing uncertainty or marking the region as unreadable, VLMs tend to produce textual predictions that are unsupported by visual evidence. The fact that different models generate different words for the same region further suggests that these predictions arise from language priors rather than from information present in the image.

Another particularly concerning example of this behavior was encountered in lists of names. Even when textual content is clear, VLMs may generate additional names that are not present in the original document, as shown in Fig.~\ref{fig:names}. This directly affects named entities, which are often the primary target of downstream applications such as information extraction and knowledge graph construction. Even small perturbations, such as introducing non-existent names, can lead to significant semantic distortions in the extracted information.

\subsection{Semantic Substitutions and Insertions}
\label{sec:substitutions}

\begin{figure}[t!]
\centering
\includegraphics[width=\linewidth]{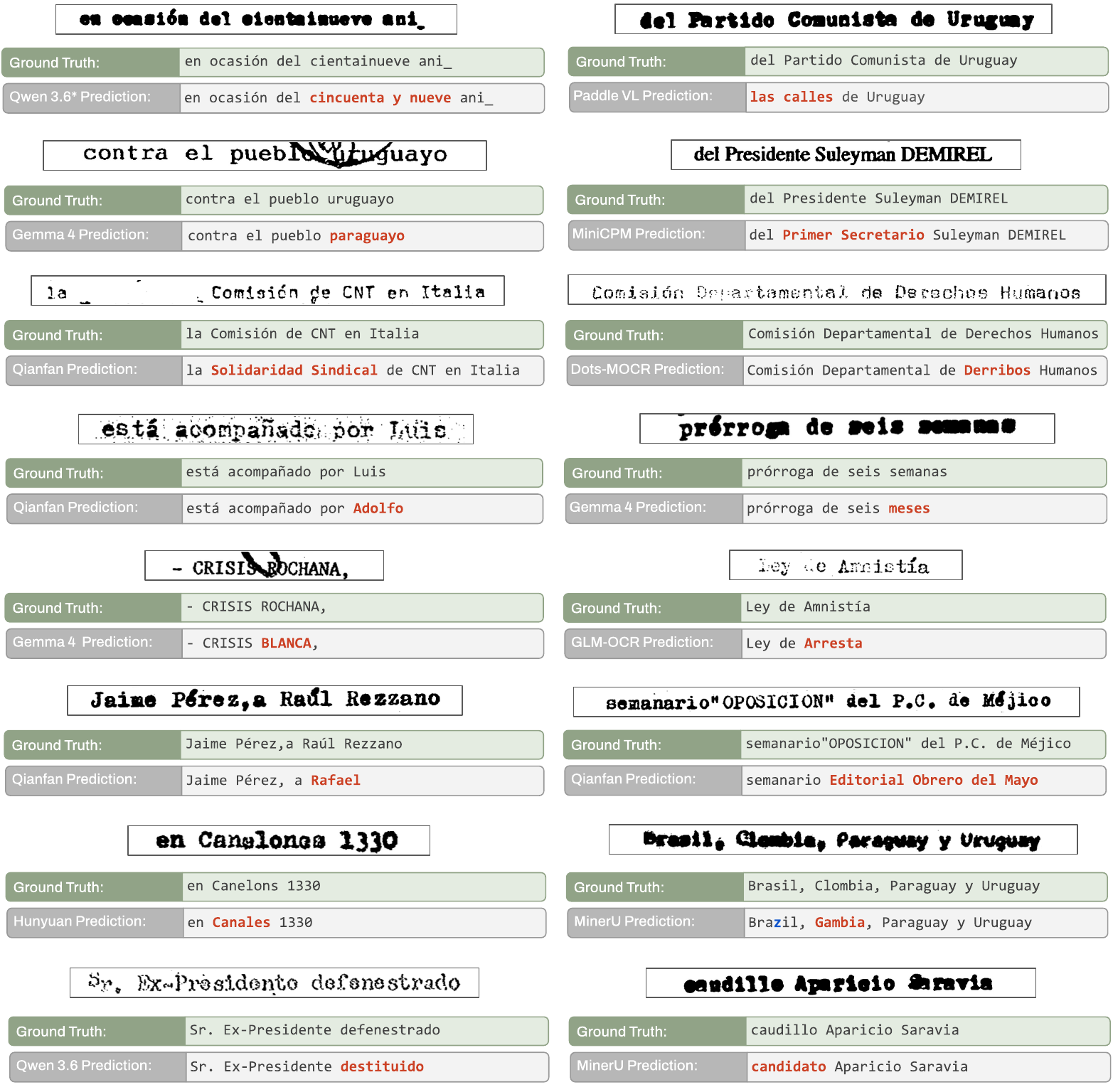}
\caption{Examples of semantic substitutions and insertions, highlighted in red. The model replaces words or entities with alternative expressions that are often syntactically plausible but semantically incorrect. Other errors are highlighted in blue.}
\label{fig:semantic}
\end{figure}

Semantic substitutions arise when the model replaces a word or expression with another that alters the meaning of the original text. Several examples of this failure mode are shown in Fig.~\ref{fig:semantic}. These cases reveal a consistent pattern that goes beyond traditional OCR errors: across diverse document types, the models often preserve the overall syntactic structure of the input while selectively modifying or replacing specific words and expressions.

These modifications are not random character-level mistakes but semantically coherent substitutions. Importantly, in these examples, they tend to preserve the entity type (e.g., person, organization, or country) while altering the underlying identity. For instance, personal names are replaced with those of different individuals, and countries are swapped in ways that remain linguistically natural. This indicates that the model is relying on learned priors over entity distributions. A particularly critical aspect of this behavior is the resulting violation of \emph{named entity integrity}.

In most cases, the image regions corresponding to these errors are clearly legible, suggesting that the failures are not driven by visual degradation or ambiguity, but by the model deviating from the source content during generation. Despite these deviations, the predicted text remains fully fluent and readable, creating a misleading impression of correctness under superficial evaluation.

These substitutions usually remain visually and linguistically close to the ground truth, leading to only modest increases in CER or WER despite substantial semantic divergence. This discrepancy is particularly concerning in historical and legal archives, where correctness depends on factual fidelity rather than string similarity. Overall, this highlights a fundamental limitation of standard OCR evaluation metrics, which may significantly underestimate errors that preserve structure while altering meaning.

\section{Conclusions and Discussion}
\label{sec:conclusions}

In this work, we presented a comparative evaluation of traditional OCR systems and recent Vision-Language Models (VLMs) on the Berrutti dataset, a challenging collection of historical documents derived from microfilm scans of the Uruguayan dictatorship period. Using standard OCR metrics, namely Character Error Rate (CER) and Word Error Rate (WER), we observed that VLM-based approaches consistently outperform traditional OCR pipelines, achieving substantially lower error rates. However, our qualitative analysis reveals that this quantitative improvement does not fully capture the nature of the errors produced by modern VLMs. While traditional OCR systems primarily fail through recognition errors, VLM-based systems exhibit qualitatively different failure modes. Importantly, many of these errors preserve fluency and syntactic structure while altering the underlying factual content.

This discrepancy highlights a fundamental limitation of standard OCR evaluation protocols: CER and WER implicitly assume that all errors are equivalent, regardless of their semantic significance. Our results show that this assumption is increasingly inadequate in the context of generative OCR systems. In particular, errors affecting named entities can have disproportionately large implications for archival use, despite contributing only marginally to the aggregate metrics.

These observations suggest that OCR evaluation should move beyond purely string-based similarity measures and incorporate more structured notions of correctness. A promising direction is the introduction of target-based evaluation protocols, in which specific information units, such as named entities, temporal expressions, identifiers, and key factual elements, are explicitly annotated and evaluated. Such a framework would allow to distinguish between minor orthographic deviations and semantically critical errors, which is particularly relevant in historical and legal archives.

Beyond evaluation, our findings raise questions about the reliability of VLM-based OCR in downstream applications. Prior work has shown that transcription errors can significantly impact tasks such as information retrieval and named entity recognition. However, the error profiles observed in VLMs differ from those of traditional OCR systems, suggesting that existing analyses may not fully transfer to this new setting. Understanding how the reported errors propagate through document processing pipelines remains an important open problem.

Another relevant direction concerns the notion of confidence and uncertainty estimation in VLM-based transcription. Unlike traditional OCR systems, VLMs generate text using an underlying probability model which could be used to produce an estimate of uncertainty in their decision. Exploring whether hallucinated or semantically incorrect outputs correlate with measurable uncertainty could enable the development of filtering or correction strategies.

Finally, although more experimentation is needed to confirm this, results such as those in Figure~\ref{fig:semantic} hint that classical OCR pipelines and VLM-based systems may fail in different places, under different circumstances. If this were confirmed, then there may be room for improvement by combining the output of several OCRs in one hybrid system.

Overall, this study highlights that improvements in standard OCR benchmarks do not necessarily translate into improved transcription fidelity in a broader sense. As OCR systems increasingly transition toward generative multimodal models, we argue that evaluation protocols must evolve accordingly to better reflect the requirements of historical, legal, and archival applications.

\section*{Acknowledgments}

The research that originated the results presented in this publication was partly supported by the Agencia Nacional de Investigación e Innovación of Uruguay, and by the France 2030 CollabNext project. The experiments presented in this paper used ClusterUY \cite{clusterUy}.

%
%
\bibliographystyle{splncs04}

\begin{thebibliography}{10}
\providecommand{\url}[1]{\texttt{#1}}
\providecommand{\urlprefix}{URL }
\providecommand{\doi}[1]{https://doi.org/#1}

\bibitem{low-resource}
Agarwal, M., Anastasopoulos, A.: A concise survey of {OCR} for low-resource
  languages. In: Proceedings of the 4th Workshop on Natural Language Processing
  for Indigenous Languages of the Americas. pp. 88--102. Association for
  Computational Linguistics, Mexico (Jun 2024)

\bibitem{qwen3_6_alibaba}
{Alibaba Cloud}, Team, Q.: Qwen3.6 series: Qwen3.6-35b-a3b model card.
  \url{https://huggingface.co/Qwen/Qwen3.6-35B-A3B} (2026), accessed:
  2026-06-07

\bibitem{barrere}
Barrere, K., Soullard, Y., Lemaitre, A., Co{\"u}asnon, B.: Training transformer
  architectures on few annotated data: An application to historical handwritten
  text recognition. International Journal on Document Analysis and Recognition
  \textbf{27}(4),  553--566 (2024). \doi{10.1007/s10032-023-00459-2}

\bibitem{belzarena2025improving}
Belzarena, D., Mowlavi, S., Artola, A., Mari{\~n}o, C., Gardella, M.,
  Ram{\'i}rez, I., Tadros, A., He, R., Bottaioli, N., Rajaei, B., Randall, G.,
  Morel, J.M.: Improving ocr using internal document redundancy. In: Document
  Analysis and Recognition -- ICDAR 2025. Lecture Notes in Computer Science,
  vol. 16026, pp. 244--260. Springer (2025). \doi{10.1007/978-3-032-04627-7_14}

\bibitem{metatr}
Boillet, M., Tarride, S., Kermorvant, C.: Metatr: A multilingual, evolving
  benchmark for automatic text recognition. In: Fink, G.A., Forn{\'e}s, A.,
  Kise, K., Lopresti, D. (eds.) Document Analysis and Recognition -- ICDAR
  2026. pp. 20--36. Springer Nature Switzerland, Cham (2027)

\bibitem{ocropus}
Breuel, T.M.: {The OCRopus open source OCR system}. In: Yanikoglu, B.A.,
  Berkner, K. (eds.) Document Recognition and Retrieval XV. vol.~6815, p.
  68150F. International Society for Optics and Photonics, SPIE (2008)

\bibitem{caetano2011}
Caetano, G.: Los archivos represivos en los procesos de "justicia
  transicional": una cuestión de derechos. Perfiles Latinoamericanos
  \textbf{19}(37),  9–32 (ene 2011)

\bibitem{caetano2017}
Caetano, G.: Los archivos represivos y el debate sobre los criterios para su
  mejor utilizaci{\'o}n como instrumento de justicia y de derechos. Claves.
  Revista de Historia  \textbf{3}(5),  155--183 (2017)

\bibitem{escriptorium}
Chagu{\'e}, A.: escriptorium : une application libre pour la transcription
  automatique des manuscrits. Arabesques  \textbf{107}, ~25 (2022)

\bibitem{transkribus}
Colutto, S., Kahle, P., Hackl, G., Muehlberger, G.: Transkribus. a platform for
  automated text recognition and searching of historical documents. In: 2019
  15th International Conference on eScience (eScience). pp. 463--466 (2019).
  \doi{10.1109/eScience.2019.00060}

\bibitem{paddleocrvl15multitask09bvlm}
Cui, C., Sun, T., Liang, S., Gao, T., Zhang, Z., Liu, J., Wang, X., Zhou, C.,
  Liu, H., Lin, M., Zhang, Y., Zhang, Y., Liu, Y., Yu, D., Ma, Y.:
  Paddleocr-vl-1.5: Towards a multi-task 0.9b vlm for robust in-the-wild
  document parsing (2026), \url{https://arxiv.org/abs/2601.21957}

\bibitem{ppocrv5}
Cui, C., Sun, T., Lin, M., Gao, T., Zhang, Y., Liu, J., Wang, X., Zhang, Z.,
  Zhou, C., Liu, H., Zhang, Y., Lv, W., Huang, K., Zhang, Y., Zhang, J., Zhang,
  J., Liu, Y., Yu, D., Ma, Y.: Paddleocr 3.0 technical report (2025),
  \url{https://arxiv.org/abs/2507.05595}

\bibitem{chandra}
Datalab: Chandra ocr 2. \url{https://github.com/datalab-to/chandra} (2026),
  gitHub repository

\bibitem{qianfan}
Dong, D., Zheng, M., Xu, D., Luo, C., Zhuang, B., Li, Y., He, R., Wang, H.,
  Zhang, W., Wang, W., Wang, Y., Xiong, X., Zheng, A., Zuo, X., Ou, Z., Gu, J.,
  Guo, Q., Wu, J., Yin, D., Shen, D.: Qianfan-ocr: A unified end-to-end model
  for document intelligence (2026), \url{https://arxiv.org/abs/2603.13398}

\bibitem{duan2026glmocrtechnicalreport}
Duan, S., Xue, Y., Wang, W., Su, Z., Liu, H., Yang, S., Gan, G., Wang, G.,
  Wang, Z., Yan, S., Jin, D., Zhang, Y., Wen, G., Wang, Y., Zhang, Y., Zhang,
  X., Hong, W., Cen, Y., Yin, D., Chen, B., Yu, W., Gu, X., Tang, J.: Glm-ocr
  technical report (2026), \url{https://arxiv.org/abs/2603.10910}

\bibitem{etcheverry2021}
Etcheverry, L., Agorio, L., Bacigalupe, V., Barreiro, S., Bing, E., Blixen, S.,
  Calegari, D., Cardozo, L., Carpani, F., Chavat, F., Garat, D., Gomez, A.,
  Fernández, E., Fioritto, F., Hernandez~Muñiz, F., Rosa, A., Stabile, J.,
  Tiscornia, J., Patiño, N., Randall, G.: A computational framework for the
  analysis of the uruguayan dictatorship archives. In: Proceedings of Qurator
  2021 (02 2021)

\bibitem{google_gemma4_docs}
Google: Gemma 4 model overview. \url{https://ai.google.dev/gemma/docs/core}
  (2026), accessed: 2026-06-07

\bibitem{ctc}
Graves, A., Fern{\'a}ndez, S., Gomez, F., Schmidhuber, J.: Connectionist
  temporal classification: Labelling unsegmented sequence data with recurrent
  neural networks. In: Proceedings of the 23rd International Conference on
  Machine Learning. pp. 369--376. Association for Computing Machinery (2006).
  \doi{10.1145/1143844.1143891}

\bibitem{ocrNERlinking}
Hamdi, A., Linhares~Pontes, E., Sidere, N., Coustaty, M., Doucet, A.: In-depth
  analysis of the impact of ocr errors on named entity recognition and linking.
  Natural Language Engineering  \textbf{29}(2),  425--448 (2023).
  \doi{10.1017/S1351324922000110}

\bibitem{whenPostCorrectionNER}
Huynh, V.N., Hamdi, A., Doucet, A.: When to use ocr post-correction for named
  entity recognition? In: Digital Libraries at Times of Massive Societal
  Transition, pp. 33--42. Springer (2020). \doi{10.1007/978-3-030-64452-9_3}

\bibitem{easyocr}
JaidedAI: Easyocr. \url{https://github.com/jaidedai/easyocr} (2024), gitHub
  repository

\bibitem{BeyondCERWER}
Jaud, A., Hamdi, A., Doucet, A., Jatowt, A., Coustaty, M.: Beyond cer and wer:
  How does ocr really impact information retrieval? In: 2025 ACM/IEEE Joint
  Conference on Digital Libraries (JCDL). pp. 30--39 (2025).
  \doi{10.1109/JCDL67857.2025.00014}

\bibitem{kang2020}
Kang, L., Rusinol, M., Fornes, A., Riba, P., Villegas, M.: Unsupervised writer
  adaptation for synthetic-to-real handwritten word recognition. In:
  Proceedings of the IEEE/CVF Winter Conference on Applications of Computer
  Vision (WACV) (March 2020)

\bibitem{kraken}
Kiessling, B.: {The Kraken OCR system} (Aug 2025), \url{https://kraken.re}

\bibitem{challenge2026icdar}
Kiessling, B., Boutreux, A., Caers, B., Levenson, M.G., Kestemont, M.,
  Michalcov{\'a}, A., Pinche, A., Vandyck, C., Vlachou~Efstathiou, M.,
  Cl{\'e}rice, T.: Icdar 2026 competition on multilingual medieval handwriting
  recognition. In: Document Analysis and Recognition -- ICDAR 2026. pp.
  353--369. Springer (2026). \doi{10.1007/978-3-032-36042-7_21}

\bibitem{lessa2013}
Lessa, F.: Memory and Transitional Justice in Argentina and Uruguay: Against
  Impunity. Palgrave Macmillan, New York (2013)

\bibitem{lessa2022}
Lessa, F.: The Condor Trials: Transnational Repression and Human Rights in
  South America. Yale University Press (2022)

\bibitem{dotsOCR}
Li, Y., Yang, G., Liu, H., Wang, B., Zhang, C.: dots.ocr: Multilingual document
  layout parsing in a single vision-language model (2025),
  \url{https://arxiv.org/abs/2512.02498}

\bibitem{ocrbench}
Liu, Y., Li, Z., Huang, M., Yang, B., Yu, W., Li, C., Yin, X.C., Liu, C.L.,
  Jin, L., Bai, X.: Ocrbench: On the hidden mystery of ocr in large multimodal
  models. Science China Information Sciences  \textbf{67}(12) (2024).
  \doi{10.1007/s11432-024-4235-6}

\bibitem{doctr2021}
Mindee: doctr: Document text recognition. \url{https://github.com/mindee/doctr}
  (2021)

\bibitem{clusterUy}
Nesmachnow, S., Iturriaga, S.: {Cluster-UY}: Collaborative scientific high
  performance computing in {U}ruguay. In: Torres, M., Klapp, J. (eds.)
  Supercomputing. pp. 188--202 (2019), \url{https://cluster.uy/}

\bibitem{ocr-d}
Neudecker, C., Baierer, K., Federbusch, M., Boenig, M., W\"{u}rzner, K.M.,
  Hartmann, V., Herrmann, E.: Ocr-d: An end-to-end open source ocr framework
  for historical printed documents. In: Proceedings of the 3rd International
  Conference on Digital Access to Textual Cultural Heritage. p. 53–58.
  Association for Computing Machinery, NY, USA (2019)

\bibitem{omnidocbench}
Ouyang, L., Qu, Y., Zhou, H., Zhu, J., Zhang, R., Lin, Q., Wang, B., Zhao, Z.,
  Jiang, M., Zhao, X., Shi, J., Wu, F., Chu, P., Liu, M., Li, Z., Xu, C.,
  Zhang, B., Shi, B., Tu, Z., He, C.: Omnidocbench: Benchmarking diverse pdf
  document parsing with comprehensive annotations (2025),
  \url{https://arxiv.org/abs/2412.07626}

\bibitem{paruchuri2025surya}
Paruchuri, V., Team, D.: Surya: A lightweight document ocr and analysis
  toolkit. \url{https://github.com/datalab-to/surya} (2025), gitHub repository

\bibitem{pippi2023}
Pippi, V., Cascianelli, S., Kermorvant, C., Cucchiara, R.: How to choose
  pretrained handwriting recognition models for single writer fine-tuning. In:
  Document Analysis and Recognition -- ICDAR 2023. pp. 330--347. Lecture Notes
  in Computer Science, Springer (2023). \doi{10.1007/978-3-031-41679-8_19}

\bibitem{olmbench}
Poznanski, J., Rangapur, A., Borchardt, J., Dunkelberger, J., Huff, R., Lin,
  D., Rangapur, A., Wilhelm, C., Lo, K., Soldaini, L.: olmocr: Unlocking
  trillions of tokens in pdfs with vision language models (2025),
  \url{https://arxiv.org/abs/2502.18443}

\bibitem{puigcerver}
Puigcerver, J.: Are multidimensional recurrent layers really necessary for
  handwritten text recognition? In: 2017 14th IAPR International Conference on
  Document Analysis and Recognition (ICDAR). vol.~01, pp. 67--72 (2017).
  \doi{10.1109/ICDAR.2017.20}

\bibitem{ocr4all}
Reul, C., Christ, D., Hartelt, A., Balbach, N., Wehner, M., Springmann, U.,
  Wick, C., Grundig, C., B{\"u}ttner, A., Puppe, F.: Ocr4all---an open-source
  tool providing a (semi-)automatic ocr workflow for historical printings.
  Applied Sciences  \textbf{9}(22), ~4853 (2019). \doi{10.3390/app9224853}

\bibitem{churro}
Semnani, S.J., Zhang, H., He, X., Tekgürler, M., Lam, M.S.: Churro: Making
  history readable with an open-weight large vision-language model for
  high-accuracy, low-cost historical text recognition (2025),
  \url{https://arxiv.org/abs/2509.19768}

\bibitem{crnn}
Shi, B., Bai, X., Yao, C.: An end-to-end trainable neural network for
  image-based sequence recognition and its application to scene text
  recognition. IEEE Transactions on Pattern Analysis and Machine Intelligence
  \textbf{39}(11),  2298--2304 (2017). \doi{10.1109/TPAMI.2016.2646371}

\bibitem{simon2026}
Simon, T., Tranouez, P., Nicolas, S., Chatelain, C., Paquet, T.: Few-shot
  writer adaptation via multimodal in-context learning. In: Fink, G.A.,
  Forn{\'e}s, A., Kise, K., Lopresti, D. (eds.) Document Analysis and
  Recognition -- ICDAR 2026. pp. 608--625. Springer Nature Switzerland, Cham
  (2027). \doi{10.1007/978-3-032-36033-5_36}

\bibitem{tesseract}
Smith, R.: An overview of the tesseract ocr engine. In: Ninth International
  Conference on Document Analysis and Recognition. vol.~2, pp. 629--633 (2007).
  \doi{10.1109/ICDAR.2007.4376991}

\bibitem{assessingOCRNLP}
van Strien, D., Beelen, K., Ardanuy, M., Hosseini, K., McGillivray, B.,
  Colavizza, G.: Assessing the impact of ocr quality on downstream nlp tasks.
  In: Proceedings of the 12th International Conference on Agents and Artificial
  Intelligence -- Volume 1: ARTIDIGH. pp. 484--496. SciTePress (2020).
  \doi{10.5220/0009169004840496}

\bibitem{hunyuanvisionteam2025hunyuanocrtechnicalreport}
Team, H.V., Lyu, P., Wan, X., Li, G., Peng, S., Wang, W., Wu, L., Shen, H.,
  Zhou, Y., Tang, C., Yang, Q., Peng, Q., Luo, B., Yang, H., Zhang, X., Zhang,
  J., Peng, H., Yang, H., Xie, S., Zhou, L., Pei, G., Wu, B., Wu, K., Yang, J.,
  Wang, B., Liu, K., Zhu, J., Jiang, J., Linus, Hu, H., Zhang, C.: Hunyuanocr
  technical report (2025), \url{https://arxiv.org/abs/2511.19575}

\bibitem{TransitionalJustice}
Teitel, R.G.: Transitional Justice. Oxford University Press (2000)

\bibitem{mineru25p}
Wang, B., He, T., Ouyang, L., Wu, F., Zhao, Z., Chu, T., Qu, Y., Jin, Z., Zeng,
  W., Miao, Z., Xu, B., Niu, J., Cai, M., Qiu, J., Zhang, Q., Ma, D., Sun, Y.,
  Dong, H., Zhang, W., Xiao, J., Shi, J., Liao, P., Zhao, X., Zhong, H., Wei,
  L., Yu, J., Yang, J., Li, W., Wang, S., Wu, Q., Zhou, X., Li, W., Li, Z., Tu,
  Z., Wu, J., Wu, L., Xu, C., Chen, K., Zhang, W., Qiao, Y., Zhou, B., Lin, D.,
  He, C.: Mineru2.5-pro: Pushing the limits of data-centric document parsing at
  scale (2026), \url{https://arxiv.org/abs/2604.04771}

\bibitem{wei2025deepseek}
Wei, H., Sun, Y., Li, Y.: Deepseek-ocr: Contexts optical compression. arXiv
  preprint arXiv:2510.18234  (2025)

\bibitem{calamari}
Wick, C., Reul, C., Puppe, F.: Calamari -- a high-performance tensorflow-based
  deep learning package for optical character recognition. Digital Humanities
  Quarterly  \textbf{14}(2) (2020). \doi{10.63744/wnzk3f3ymuez}

\bibitem{yu2025minicpmv45}
Yu, T., Wang, Z., Wang, C., Huang, F., Ma, W., He, Z., Cai, T., Chen, W.,
  Huang, Y., Zhao, Y., Xu, B., Cui, J., Xu, Y., Ruan, L., Zhang, L., Liu, H.,
  Tang, J., Liu, H., Guo, Q., Hu, W., He, B., Zhou, J., Cai, J., Qi, J., Guo,
  Z., Chen, C., Zeng, G., Li, Y., Cui, G., Ding, N., Han, X., Yao, Y., Liu, Z.,
  Sun, M.: Minicpm-v 4.5: Cooking efficient mllms via architecture, data, and
  training recipe (2025), \url{https://arxiv.org/abs/2509.18154}

\bibitem{zhang2026democratizingmedievalenglishlegal}
Zhang, M., Wang, E., Whatley, C., Strickland, S., Bannon, D.: Democratizing
  the medieval english legal tradition. In: Fink, G.A., Forn{\'e}s, A., Kise,
  K., Lopresti, D. (eds.) Document Analysis and Recognition -- ICDAR 2026. pp.
  378--394. Springer Nature Switzerland, Cham (2027). \doi{10.1007/978-3-032-36023-6_22}

\bibitem{dotsMOCR}
Zheng, H., Li, Y., Zhang, K., Xin, L., Zhao, G., Liu, H., Chen, J., Lou, J.,
  Qiu, J., Fu, Q., Yang, R., Jiang, S., Luo, W., Su, W., Zhang, W., Zhu, X.,
  Li, Y., ma, Y., Chen, Y., Yu, Z., Yang, G., Zhang, C., Zhang, L., Liu, Y.,
  Bai, X.: Multimodal ocr: Parse anything from documents (2026),
  \url{https://arxiv.org/abs/2603.13032}

\end{thebibliography}

%
%
\end{document}